%% file: aa.tex
\documentclass[11pt,a4paper]{article}
\usepackage[hyperref]{emnlp-ijcnlp-2019}
\usepackage{times}
\usepackage{latexsym}
\usepackage{amsmath}
\usepackage{amssymb}
\usepackage{color}
\usepackage{xspace}
\usepackage{multirow}
\usepackage{url}
\usepackage{graphicx}
\usepackage{tikz}
\usepackage{subcaption}
\usepackage{breakurl}
\usetikzlibrary{patterns}

\newcommand\percent{\hspace*{-0.4ex}\ensuremath{^{^{[\%]}}}}
\newcommand\BLEU{\textsc{Bleu}\xspace}
\newcommand\BLEUpercent{\BLEU\percent\xspace}
\newcommand\AER{\textsc{Aer}\xspace}
\newcommand\AERpercent{\AER\percent\xspace}
\newcommand\GIZA{\textsc{Giza++}\xspace}

\aclfinalcopy 

\title{Jointly Learning to Align and Translate with Transformer Models}

\author{Sarthak Garg\quad Stephan Peitz\quad  Udhyakumar Nallasamy\quad  Matthias Paulik\\
  Apple Inc. \\
  {\tt \{sarthak\_garg, speitz, udhay, mpaulik\}@apple.com}
}

\date{}

\begin{document}
\maketitle
\begin{abstract}
The state of the art in machine translation (MT) is governed by neural approaches, which typically provide superior translation accuracy over statistical approaches. However, on the closely related task of word alignment, traditional statistical word alignment models often remain the go-to solution. 
In this paper, we present an approach to train a Transformer model to produce both accurate translations and alignments.
We extract discrete alignments from the attention probabilities learnt during regular neural machine translation model training and leverage them in a multi-task framework to optimize towards translation and alignment objectives.
We demonstrate that our approach produces competitive results compared to GIZA++ trained IBM alignment models without sacrificing translation accuracy and outperforms previous attempts on Transformer model based word alignment. Finally, by incorporating IBM model alignments into our multi-task training, we report significantly better alignment accuracies compared to
GIZA++ on three publicly available data sets.
Our implementation has been open-sourced\footnote{Code can be found at \url{https://github.com/pytorch/fairseq/pull/1095}}.
\end{abstract}

\section{Introduction}
Neural machine translation (NMT) constitutes the state of the art in MT, with the Transformer model architecture~\cite{vaswani2017transformer} beating other neural architectures in competitive MT evaluations. The attention mechanism used in NMT models was motivated by the need to model word alignments, however it is now well known that the attention probabilities can differ significantly from word alignments in the traditional sense~\cite{koehn2017six}, since attending to the context words rather than the aligned source words might be helpful for translation. The presence of multi-layer, multi-head attention mechanisms in the Transformer model further complicates interpreting the attention probabilities and extracting high quality discrete alignments from them.

Finding source to target word alignments has many applications in the context of MT. A straightforward application of word alignments is to generate bilingual lexica from parallel corpora. Word alignments have also been used for 
external dictionary assisted translation~\cite{chatterjee2017guiding, alkhouli18:transformeralign, arthur2016incorporating} to improve translation of low frequency words or to comply with certain terminology guidelines. Documents and webpages often contain word annotations such as formatting styles and hyperlinks, which need to be preserved in the translation. In such cases, word alignments can be used to transfer these annotations from the source sentence to its translation. In user facing
translation services, providing word alignments as additional information to the users might improve their trust and confidence, and also help them to diagnose problems such as under-translation~\cite{tu2016modeling}. 

In this work, we introduce an approach that teaches Transformer models to produce translations and interpretable alignments simultaneously:
\begin{itemize}
\item We use a multi-task loss function combining negative log likelihood (NLL) loss used in regular NMT model training and an alignment loss supervising one attention head to learn alignments (Section \ref{sec:multitask}).
\item Conditioning on past target context is essential for maintaining the auto-regressive property for translation but can be limiting for alignment. We alleviate this problem by conditioning the different 
components of our multi-task objective on different amounts of context (Section \ref{sec:fullcontext}). 
\item We demonstrate that the system can be supervised using seed alignments obtained by carefully averaging the attention probabilities of a regular NMT model (Section \ref{sec:avg}) or alignments obtained 
from statistical alignment tools (Section \ref{sec:external}) 
\end{itemize}    
We show that our model outperforms previous neural approaches \cite{peter2017:Foresight, Zenkel2019} and statistical alignment models \cite{giza2003} in terms of alignment accuracy without suffering any degradation of translation accuracy.
\section{Preliminaries}
\subsection{Word Alignment Task}
Given a sentence $f_{1}^{J}= f_1, \ldots, f_j, \ldots f_J$ in the source language \textit{and} its translation $e_1^I = e_1, \ldots, e_i, \ldots  e_I$ in the target language, an alignment $\mathcal{A}$ 
is defined as a subset of the Cartesian product of the word positions \cite{giza2003}.
\begin{equation}
\mathcal{A} \subseteq \{(j, i): j = 1,\ldots,J;i=1,\ldots,I\}
\end{equation}
The word alignment task aims to find a discrete alignment representing a many-to-many mapping from the source words to their corresponding translations in the target sentence.
\subsection{Transformer Model}
\label{sec:transformer}
The Transformer model~\cite{vaswani2017transformer} is an encoder-decoder model that only relies on attention for computing the contextual representations for source and target sentences.
Both the encoder and decoder are composed of multiple layers, each of which includes a multi-head self-attention and a feed-forward sub-layer.
Layers in the decoder additionally apply a multi-head encoder-decoder attention between the self-attention and the feed-forward sub-layers.
To maintain the auto-regressive property, the self-attention sub-layer in the decoder attends to the representations of only the \textit{past} tokens computed by the lower layer.

In this work, we will be focusing on guiding the encoder-decoder attention sub-layer in the decoder. Let $d_{emb}, d_{k}, d_{v}, N$ denote the embedding dimension, dimensions of the key and value projections and number of heads, respectively.
As described in \newcite{vaswani2017transformer}, for this sub-layer, the output of the previous 
decoder layer corresponding to the $i^{th}$ target token is used as a query vector $\mathbf{q}^{i} \in \mathbb{R}^{1 \times d_{emb}}$ and the encoder output for all the source tokens are packed together as the value $V \in \mathbb{R}^{J \times d_{emb}}$ and key $K \in \mathbb{R}^{J \times d_{emb}}$ matrices.
To compute the output $\mathcal{M}(\mathbf{q}^{i}, K, V)$, $N$ heads first project the query vector and the key and value matrices into different subspaces, compute attention in their own subspaces, aggregate their outputs and project back to the original space:
\begin{align}
& \mathbf{\tilde{q}}^{i}_{n} = \mathbf{q}^{i}W_{n}^{Q}, \tilde{K}_{n} = KW_{n}^{K}, \tilde{V}_{n} = VW_{n}^{V} \\
&H^{i}_{n} = \text{Attention}(\mathbf{\tilde{q}}^{i}_{n}, \tilde{K}_{n}, \tilde{V}_{n}) \\
&\mathcal{M}(\mathbf{q}^{i}, K, V) = \text{Concat}(H_{1}^{i}, \ldots , H_{N}^{i})W^{O}, 
\end{align}
where the projection matrices $W_n^{Q}$, $W_{n}^{K}$, $W_{n}^{V}$ and $W^{O}$ are learnt parameters of the $n^{th}$ head.
Each head employs a scaled dot-product attention:
\begin{align}
&\text{Attention}(\mathbf{\tilde{q}}^{i}_{n},\tilde{K}_{n}, \tilde{V}_{n}) = \mathbf{a}^{i}_{n}\tilde{V}_{n}, \\
&\text{where }\mathbf{a}^{i}_{n} = \text{softmax}(\frac{\mathbf{\tilde{q}}^{i}_{n}\tilde{K}^{T}_{n}}{\sqrt{d_{k}}}).
\end{align} 
The vector $\mathbf{a}^{i}_{n} \in \mathbb{R}^{1 \times J}$ denotes the attention probabilities for the $i^{th}$ target token over all the source tokens, computed by the $n^{th}$ attention head.
For any particular head, an attention matrix $A_{I \times J}$ can be constructed by grouping together the vectors $\mathbf{a}^{i}_{n}$ corresponding to all the target tokens.
In the following sections, we analyze the quality of alignments that can be extracted from these attention matrices $A_{I \times J}$ and describe how they can be effectively supervised to learn 
word alignments.

\section{Baseline Methods}
A common baseline approach to extract word alignments from a regular NMT trained Transformer, is to average over all attention matrices $A_{I \times J}$ computed across all layers and heads.
The resulting matrix gives a probability distribution over all source tokens for each target token. This distribution is 
then converted to a discrete alignment by aligning each target word to the corresponding source word with the highest attention probability.

\newcite{peter2017:Foresight} guide the attention probabilities to be close to the alignments obtained from statistical MT toolkits by imposing an additional loss based on the
distance between the alignment and attention distributions. They get improvements in alignment accuracy over previous works based on guided alignment training by feeding the \textit{current} target word to the attention module, providing it more context about the target sentence.

\newcite{Zenkel2019} proposed an method that does not rely on alignments from external toolkits for training. They instead add an extra attention layer on top of the Transformer architecture and directly optimize its activations towards predicting the given target word. 

All the above methods involve training models for both the directions to get bidirectional alignments. These bidirectional alignments are then merged using
 the {\tt grow diagonal} heuristic~\cite{sym2005}. 
\section{Proposed Method}
\subsection{Averaging Layer-wise Attention Scores}
\label{sec:avg}
The attention heads in a single layer are symmetrical, but the different layers themselves can learn drastically different alignments. To better understand the behavior of the encoder-decoder 
attention learnt at different layers, we average the attention matrices computed across all heads within each layer and evaluate the obtained alignments. 
We show that the attention probabilities from the penultimate layer naturally tend to learn alignments and provide significantly better results compared to naively 
averaging across all layers (cf.~Section~\ref{sec:resultsAverage}). For the rest of the paper, we refer to the former method as the layer average baseline.
\subsection{Multi-task Learning}
\label{sec:multitask}
Translation and alignment tasks are very closely related. NMT models with attention~\cite{bahdanau2014neural} have also shown to learn alignments in the intermediate attention layer. A neural model receiving supervision from given translations \textit{and} given alignments
can therefore benefit from multi-task learning by exploiting the correlations between these two tasks. 

Annotating word alignments is a laborious and expensive task, but the layer average baseline described in Section~\ref{sec:avg} is able to generate reasonably good alignments in an unsupervised 
manner. We thus use the alignments generated by the layer average baseline as labels for supervising our model.
We first convert the alignments into a probability distribution over source tokens for every target token. Let $G_{I \times J}$ denote a $0$-$1$ matrix such that $G_{i, j} = 1$
if the $j^{th}$ source token is aligned to the $i^{th}$ target token. We simply normalize the rows of matrix $G$ corresponding to target tokens that are aligned to at least one
source token to get a matrix $G^{p}$.
As described in Section~\ref{sec:transformer}, the Transformer model computes multiple attention probability distributions over source tokens for every target token across different heads and layers of the network. 
Since we observed that the attention probabilities from the penultimate layer most naturally tend to learn alignments (Section~\ref{sec:resultsAverage}), we arbitrarily select one head from the 
penultimate layer (subsequently referred to as the alignment head) and supervise its attention probability distribution to be close to the labeled alignment distribution ($G^{p}$). 
Let $A_{I \times J}$ denote the attention matrix computed by the alignment head. For every target word $i$, we minimize the Kullback-Leibler divergence between $G^{p}_{i}$ and $A_{i}$ which is 
equivalent to optimizing the following cross-entropy loss $\mathcal{L}_{a}$
\begin{equation}
\mathcal{L}_{a}(A) = -\frac{1}{I}\sum_{i = 1}^{I}\sum_{j = 1}^{J} G^{p}_{i, j} \log(A_{i, j}).
\end{equation}
The motivation behind supervising one head is that it gives the model the flexibility to either use the representation computed by the alignment head, or depend more on the representations computed by other heads.
We train our model to minimize $\mathcal{L}_{a}$ in conjunction with the standard NLL translation loss $\mathcal{L}_t$.
 The overall loss $\mathcal{L}$ is:
\begin{equation}
\label{eq:loss}
\mathcal{L} = \mathcal{L}_t + \lambda \mathcal{L}_a(A),
\end{equation}
where $\lambda$ is a hyperparameter.

\subsection{Providing Full Target Context}
\label{sec:fullcontext}
The Transformer decoder computes the probability of the next target token conditioned on the past target tokens and all source tokens.
This is implemented by masking the self attention probabilities, i.e. while computing the representation for the $i^{th}$ target token, the decoder can only self-attend to the representations of $\{1, 2 \ldots i - 1\}$ tokens from the
previous layer. This auto-regressive behavior of the decoder is crucial for the model to represent a valid probability distribution over the target sentence.
However, conditioning on just the past target tokens is limiting for the alignment task. As described in Section \ref{sec:multitask}, the alignment head is trained to model the alignment distribution for the $i^{th}$ target 
token given only the past target tokens and all source tokens. Since the alignment head does not know the identity of the next target token, it becomes difficult for it to learn this token's alignment to the source tokens.
Previous work has also identified this problem and alleviate it by feeding the target token to be aligned as an input to the module computing the alignment~\cite{peter2017:Foresight}, or forcing the 
module to predict the target token~\cite{Zenkel2019} or its properties, e.g.~POS tags~\cite{li2018target}.
Feeding the next target token assumes that we know it in advance and thus calls for separate translation and alignment models. Forcing the alignment module to predict target token's properties helps but still passes the information 
of the target token in an indirect manner. We overcome these limitations by conditioning the two components of our loss function on different amounts of context. The NLL loss $\mathcal{L}_t$ is conditioned on the past target tokens to preserve the auto-regressive property:
\begin{equation}
\mathcal{L}_t = - \frac{1}{I}\sum_{i = 1}^{I} \log(p(e_i | f_1^J, e_1^{i - 1})).
\end{equation}
However, the alignment loss $\mathcal{L}^{'}_{a}$ is now conditioned on the whole target sentence:
\begin{equation}
\mathcal{L}^{'}_{a} = \mathcal{L}_{a}(A | f_1^J, e_1^I).
\end{equation}
This is implemented by executing two forward passes of the decoder
model, one with the masking of the future target tokens for computing the NLL loss $\mathcal{L}_{t}$ and the other one with no masking for computing the alignment loss $\mathcal{L}^{'}_{a}$from the alignment head.
Although this formulation forces the network to learn representations adapting to both full and partial target context, Section \ref{sec:resultsAlignmentTranslate} shows that this approach does not degrade
the translation quality while improving the alignment accuracy.

\subsection{Alignment Training Data}
\label{sec:external}
Our method described so far does not rely on alignments from external statistical toolkits but performs self-training on alignments extracted from the layer average baseline.
However, \GIZA provides a robust method to compute accurate alignments.
If achieving better alignment accuracy is paramount, then our multi-task framework can also leverage alignments from \GIZA to produce even better alignment accuracy (Section~\ref{sec:resultsAlignment}). 
In this setting we use the \GIZA alignments  as labels instead of those obtained from the layer average baseline for supervising the alignment head.
\section{Experiments}

\subsection{Setup}
Our experiments show that our proposed approach is able to achieve state-of-the-art results in terms of alignment \textit{and} maintain the same translation performance.
In the following, we describe two setups to compare with previously established state-of-the-art results.

For all setups and models used in this work, we learn a joint source and target Byte-Pair-Encoding (BPE, \newcite{bpe2016}) with 32k merge operations.
We observe that even for statistical alignment models sub-word units are beneficial. To convert the alignments from sub-word-level back to word-level, we consider each target word as being aligned to a source word if an alignment between any of the target sub-words and source sub-words exists.

The alignment quality is evaluated by using the alignment error rate (\AER) introduced in~\cite{aer2000}.
Significance of the differences in \AER between two models is tested using a two-sided Wilcoxon signed-rank test ($\alpha=0.1\%$).

\subsubsection{Alignment Task}
\label{sec:alignsetup}
The purpose of the this task is to fairly compare with state-of-the-art results in terms of alignment quality and perform a hyperparameter search.
We use the same experimental setup as described in~\cite{Zenkel2019}. 
The authors provide pre-processing and scoring scripts\footnote{\url{https://github.com/lilt/alignment-scripts}} for three different datasets: Romanian$\to$English, English$\to$French and German$\to$English.
Training data and test data for Romanian$\to$English and English$\to$French are provided by the NAACL'03 \textit{Building and Using Parallel Texts} word alignment shared task\footnote{\url{http://web.eecs.umich.edu/~mihalcea/wpt/index.html\#resources}}~\cite{Mihalcea2003}.
The Romanian$\to$English training data are augmented by the Europarl v8 corpus increasing the amount of parallel sentences from 49k to 0.4M.
For German$\to$English we use the Europarl v7 corpus as training data and the gold alignments\footnote{\url{https://www-i6.informatik.rwth-aachen.de/goldAlignment/}} provided by \newcite{vilar2006}. The reference alignments were created by randomly selecting a subset of the Europarl v7 corpus and manually annotating them following the guidelines suggested in~\cite{giza2003}.
Data statistics are shown in Table~\ref{tab:dataStats}.
\begin{table}[!htb]
 \caption{Number of sentences for three datasets: German$\to$English (DeEn), Romanian$\to$English (RoEn) and English$\to$French (EnFr). The datasets include training data and test data with gold alignments.}
 \label{tab:dataStats}
 \begin{center}
    \begin{tabular}{|l|ccc|}
      \hline
                  & DeEn & RoEn & EnFr \\
      \hline\hline
      training    & 1.9M & 0.5k & 1.1M \\
      test        & 508  & 248  & 447  \\ \hline 
    \end{tabular}
  \end{center}
\end{table}

In all experiments for this task, we employ the {\tt base} transformer configuration with an embedding size of 512, 6 encoder and decoder layers, 8 attention heads, shared input and output embeddings~\cite{emb2017}, the standard {\tt relu} activation function and sinusoidal positional embedding.
The total number of parameters is 60M.
We train with a batch size of 2000 tokens on 8 Volta GPUs and use the validation translation loss for early stopping.
Furthermore, we use Adam optimizer~\cite{adam2018} with a learning rate of 3e-4, $\beta_1 = 0.9, \beta_2 = 0.98$, learning rate warmup over the first 4000 steps and inverse square root as learning rate scheduler.
The dropout probability is set to 0.1.
Additionally, we apply label smoothing with a factor of 0.1. 
To conveniently extract word alignments for both translation directions, we train bidirectional models, i.e. our models are able to translate and align from Romanian to English and vice versa.

\subsubsection{Align and Translate Task}
\label{sec:aligntranslatesetup}
The second setup is based on the WMT`18 English-German news translation task~\cite{bojar-etal-2018-findings}.
We apply the same corpus selection for bilingual data and model architecture as suggested by~\newcite{edunov-etal-2018-understanding}.
However, we slightly modify the preprocessing pipeline to be able to evaluate the alignment quality against the gold alignments provided by \newcite{vilar2006}.
We use all available bilingual data (Europarl v7, Common Crawl corpus, News Commentary v13 and Rapid corpus of EU press releases) excluding the ParalCrawl corpus. We remove sentences longer than 100 words and sentence pairs with a source/target length ratio exceeding 1.5.
This results in 5.2M parallel sentences.
We apply the Moses tokenizer~\cite{koehn-etal-2007-moses} without aggressive hyphen splitting and without performing HTML escaping of apostrophes and quotes.
Furthermore, we do not normalize punctuation marks.
We use newstest2012 as validation and newstest2014 as test set.

To achieve state-of-the-art translation results, all models in this setup are trained unidirectional and we change to the {\tt big} transformer configuration with an embedding size of 1024 and 16 attention heads.
The total number of parameters is 213M.
We train the layer average baseline with a batch size of 7168 tokens on 64 Volta GPUs for 30k updates and apply a learning rate of 1e-3, $\beta_1 = 0.9, \beta_2 = 0.98$.
The dropout probability is set to 0.3. All other hyperparameters are as described in the previous section.
Since training the multi-task models consumes more memory, we need to half the batch size, increase the number of updates accordingly and adapt the learning rate to 7e-4.
We average over the last 10 checkpoints and run inference with a beam size of 5.

To fairly compare against state-of-the-art translation setups, we compute \BLEU~\cite{bleu2002} with {\tt sacreBLEU}~\cite{post-2018-call}.

\subsection{Statistical Baseline}
For both setups, the statistical alignment models are computed with the multi-threaded version of the \GIZA toolkit\footnote{\url{https://github.com/moses-smt/mgiza/}} implemented by \newcite{mgiza2008}.
\GIZA estimates IBM1-5 models and a first-order hidden Markov model (HMM) as introduced in~\cite{Brown1993} and~\cite{Vogel1996}, respectively.
In particular, we perform 5 iterations of IBM1, HMM, IBM3  and IBM4.
Furthermore, the alignment models are trained in both translation directions and symmetrized by employing the {\tt grow-diagonal} heuristic~\cite{sym2005}.
We use the resulting word alignments to supervise the alignment loss for the method described in Section~\ref{sec:external}.
\subsection{Averaging Attention Results}
\label{sec:resultsAverage}
For our experiments, we use the data and Transformer model setup described in Section~\ref{sec:alignsetup}. We perform the evaluation of alignments obtained by layer wise averaging of attention probabilities as described
in Section~\ref{sec:avg}. 
\begin{table}[!htb]
 \caption{\AERpercent per layer for all three language pairs: German$\to$English (DeEn), Romanian$\to$English (RoEn) and English$\to$French (EnFr).}
 \label{tab:baseline}
 \begin{center}
    \begin{tabular}{|l|ccc|}
      \hline
      Layer        & DeEn & RoEn & EnFr \\
      \hline\hline
      1 (bottom)   & 90.0 & 92.9 & 80.7 \\
      2            & 91.0 & 93.6 & 81.7 \\
      3            & 94.5 & 92.0 & 73.4 \\
      4            & 41.2 & 37.5 & 20.5 \\
      5            & {\bf 32.6} & {\bf 33.4} & \bf{17.0} \\
      6 (top)      & 56.3 & 48.4 & 37.9 \\ \hline
      average      & 55.8 & 38.6 & 23.2 \\ \hline
    \end{tabular}
  \end{center}
\end{table}
As shown in Table~\ref{tab:baseline}, all three language pairs exhibit a very similar pattern, wherein the attentions do not seem to learn meaningful alignments in the initial
layers and show a remarkable improvement in the higher layers. This indicates that the initial layers are focusing more on learning good representations of the sentence generated by the decoder so far by self attention.
Once good contextual representations are learnt, the higher layers fetch the relevant representations from the encoder's output via the encoder-decoder attention.
However, interestingly the penultimate layer outperforms the final layer suggesting the final layer uses the alignment-based features in the penultimate layer to derive its own representation. 

\subsection{Alignment Task Results}
\label{sec:resultsAlignment}
Table~\ref{tab:resultsAlignment} compares the performance of our methods against statistical baselines and previous neural approaches. The layer average baseline provides relatively weak alignments, which are 
used for training our \textit{multi-task} model. The improvement of the multi-task approach over the layer average baseline suggests that learning to translate helps produce better alignments as well. However
still the multi-task approach falls short of the statistical and neural baselines, which have a strong advantage of having access to the full/partial target context. 
Exposing our model to the full target context gives the largest gains in terms of \AER. Note that \textit{full context} results are directly comparable to ~\newcite{Zenkel2019} since both approaches do not leverage
external knowledge from statistical models. We suspect that we are able to outperform~\newcite{Zenkel2019} because we provide the full target context instead of only the to-be aligned target word.
Finally, by supervising our model on the alignments obtained from \GIZA (\textit{\GIZA supervised}) rather than layer average baseline, we outperform \GIZA and~\newcite{peter2017:Foresight}. 

We tuned the alignment loss weight $\lambda$ (Equation~\ref{eq:loss}) using grid search on the German$\to$English dataset. 
We achieve the best results with $\lambda = 0.05$.

\begin{table}[!htb]
 \caption{Results on the alignment task (in \AERpercent). \textsuperscript{\ddag}Difference in \AER w.r.t.~\GIZA (BPE-based) is statistically significant (p\textless0.001).}
 \label{tab:resultsAlignment}
 \begin{center}
    \begin{tabular}{|l|ccc|}
      \hline
      Model                         & DeEn & RoEn & EnFr \\
      \hline\hline
      \GIZA (word-based)         & 21.4 & 27.9 & 5.9\\
      \GIZA (BPE-based)          & 18.9 & 27.0 & 5.5 \\ \hline
      Layer average baseline             & 32.6 & 33.4 & 17.0 \\
      Multi-task                 & 25.4 & 30.7 & 12.6 \\
      + full-context             & 20.2 & 26.0 & 7.7 \\
      ++ \GIZA supervised      & {\bf 16.0\textsuperscript{\ddag}} & {\bf 23.1\textsuperscript{\ddag}} & {\bf 4.6\textsuperscript{\ddag}} \\ \hline
      \newcite{peter2017:Foresight} & 19.0 & - & - \\
      \newcite{Zenkel2019}          & 21.2 & 27.6 & 10.0 \\ \hline
    \end{tabular}
  \end{center}
\end{table}

\subsection{Align and Translate Task Results}
\label{sec:resultsAlignmentTranslate}
For fair comparison of our approach to the state-of-the-art translation models, we use the setup described in Section~\ref{sec:aligntranslatesetup}. Table~\ref{tab:resultsAlignTranslate} summarizes the results on alignment and translation tasks. The layer average baseline is based on regular NMT model training, therefore ideally it should achieve the same BLEU as \newcite{edunov-etal-2018-understanding}, however we see a small drop of 0.3 BLEU points in practice which could be caused by
the slightly
different preprocessing procedure (cf.~Section~\ref{sec:aligntranslatesetup}, no aggressive hyphen splitting/no punctuation normalization). The layer average baseline performs poorly in terms of the \AER. The Precision and Recall results for the layer average baseline demonstrate the effectiveness of symmetrization. Symmetrization removes a majority of incorrect alignments and gives a high precision ($94.2\%$) but low recall ($29.6\%$). The high precision of the layer average baseline ensures that the multi-task model receives correct alignments for supervision, enabling it to get large improvements in \AER over the layer average baseline. 

\begin{table*}[!htb]
 \caption{Results on the align and translate task. Alignment quality is reported in \AER, translation quality in \BLEU. \textsuperscript{\dag}baseline (without back-translation) {\tt sacreBLEU} results were provided in \url{https://github.com/pytorch/fairseq/issues/506\#issuecomment-464411433}. \textsuperscript{\ddag}Difference in \AER w.r.t.~\GIZA (BPE-based) is statistically significant (p\textless0.001)}
 \label{tab:resultsAlignTranslate}
 \begin{center}
    \begin{tabular}{|l|ccc|cc|}
      \hline
                                      & \multicolumn{3}{c|}{\AERpercent~{\small (Precision\percent, Recall\percent)}} & \multicolumn{2}{c|}{\BLEUpercent} \\      
      Model                           & DeEn & EnDe & Symmetrized & DeEn & EnDe \\
      \hline\hline
      \GIZA (word-based)              & 21.7 {\small (85.4, 72.1}) & 24.0 {\small (85.8, 68.2)} & 22.2 {\small (93.5, 66.5}) & - & - \\
      \GIZA (BPE-based)               & 19.0 {\small (89.1, 74.2)} & 21.3 {\small (86.8, 71.9)} & 19.6 {\small (93.2, 70.6)} & - & - \\ \hline
      Layer average baseline          & 66.8 {\small (32.0, 34.6)} & 66.5 {\small (32.5, 34.7)} & 54.8 {\small (94.2, 29.6)} & 33.1 & 28.7 \\
      Multi-task                      & 31.1 {\small (67.2, 70.7)} & 32.2 {\small (66.6, 69.1)} & 25.8 {\small (88.1, 63.8)} & 33.1 & 28.5 \\
      + full-context                  & 21.2 {\small (76.9, 80.9)} & 23.5 {\small (75.0, 78.0)} & 19.5 {\small (89.5, 72.9)} & 33.2 & 28.5 \\
      ++ \GIZA supervised           & {\bf 17.5}\textsuperscript{\ddag} {\small (80.5, 84.7)} & {\bf 19.8}\textsuperscript{\ddag} {\small (78.8, 81.7)} & {\bf 16.4}\textsuperscript{\ddag} {\small (89.6, 78.2)} & 33.1 & 28.8 \\ \hline
      \newcite{edunov-etal-2018-understanding}\textsuperscript{\dag} & - & - & - & - & 29.0 \\ \hline
    \end{tabular}
  \end{center}
\end{table*}
\begin{figure*}[!ht]
\centering
\begin{subfigure}[b]{0.35\textwidth}
  \begin{tikzpicture}[thick,scale=0.35,baseline]
  \small
           \fill[black]  (1,1) rectangle (0,0);
           \fill[black]  (2,4) rectangle (1,3);
           \fill[black]  (3,3) rectangle (2,2);
           \fill[black]  (7,7) rectangle (6,6);
           \fill[black]  (8,8) rectangle (7,7);
           \fill[black]  (9,11) rectangle (8,10);
           \fill[black]  (10,13) rectangle (9,12);

        \draw (0,0.5) node[left,align=right]{Therefore};
        \draw (0,1.5) node[left,align=right]{,};
        \draw (0,2.5) node[left,align=right]{we};
        \draw (0,3.5) node[left,align=right]{have};
        \draw (0,4.5) node[left,align=right]{to};
        \draw (0,5.5) node[left,align=right]{be};
        \draw (0,6.5) node[left,align=right]{extremely};
        \draw (0,7.5) node[left,align=right]{careful};
        \draw (0,8.5) node[left,align=right]{when};
        \draw (0,9.5) node[left,align=right]{we};
        \draw (0,10.5) node[left,align=right]{analyse};
        \draw (0,11.5) node[left,align=right]{them};
        \draw (0,12.5) node[left,align=right]{.};

        \draw (0.3,-0.1) node[right, align=left,rotate=-50]{{Daher}};
        \draw (1.3,-0.1) node[right, align=left,rotate=-50]{{m{\"u}ssen}};
        \draw (2.3,-0.1) node[right, align=left,rotate=-50]{{wir}};
        \draw (3.3,-0.1) node[right, align=left,rotate=-50]{{die}};
        \draw (4.3,-0.1) node[right, align=left,rotate=-50]{{Voraussetzungen}};
        \draw (5.3,-0.1) node[right, align=left,rotate=-50]{{mit}};
        \draw (6.3,-0.1) node[right, align=left,rotate=-50]{{allergr{\"o}{\ss}ter}};
        \draw (7.3,-0.1) node[right, align=left,rotate=-50]{{Sorgfalt}};
        \draw (8.3,-0.1) node[right, align=left,rotate=-50]{{untersuchen}};
        \draw (9.3,-0.1) node[right, align=left,rotate=-50]{{.}};

        \draw[thick] (0,0) grid (10,13);
        \end{tikzpicture}
  \begin{tikzpicture}[thick,scale=0.35,baseline]
  \small
        \fill[black] (1,1) rectangle (0,0);
        \fill[black] (2,2) rectangle (1,1);
        \fill[black] (10,3) rectangle (9,2);

        \draw (0,0.5) node[left,align=right]{I};
        \draw (0,1.5) node[left,align=right]{must};
        \draw (0,2.5) node[left,align=right]{\hphantom{Th}inform};
        \draw (0,3.5) node[left,align=right]{you};

        \draw (0.3,-0.1) node[right, align=left,rotate=-50]{{Ich}};
        \draw (1.3,-0.1) node[right, align=left,rotate=-50]{{mu{\ss}}};
        \draw (2.3,-0.1) node[right, align=left,rotate=-50]{{an}};
        \draw (3.3,-0.1) node[right, align=left,rotate=-50]{{dieser}};
        \draw (4.3,-0.1) node[right, align=left,rotate=-50]{{Stelle}};
        \draw (5.3,-0.1) node[right, align=left,rotate=-50]{{das}};
        \draw (6.3,-0.1) node[right, align=left,rotate=-50]{{Haus}};
        \draw (7.3,-0.1) node[right, align=left,rotate=-50]{{dar{\"u}ber}};
        \draw (8.3,-0.1) node[right, align=left,rotate=-50]{{in}};
        \draw (9.3,-0.1) node[right, align=left,rotate=-50]{{Kenntnis}};
        \draw (10.3,-0.1) node[right, align=left,rotate=-50]{{setzen}};

        \draw[thick] (0,0) grid (11,4);
        \end{tikzpicture}
        \caption{\GIZA}
        \label{fig:giza}
\end{subfigure}
\begin{subfigure}[b]{0.26\textwidth}
  \begin{tikzpicture}[thick,scale=0.35,baseline]
  \small
           \fill[black]  (1,1) rectangle (0,0);
           \fill[black]  (2,4) rectangle (1,3);
           \fill[black]  (3,3) rectangle (2,2);
           \fill[black]  (5,12) rectangle (4,11);
           \fill[black]  (6,9) rectangle (5,8);
           \fill[black]  (7,7) rectangle (6,6);
           \fill[black]  (8,8) rectangle (7,7);
           \fill[black]  (9,11) rectangle (8,10);
           \fill[black]  (10,13) rectangle (9,12);

        \draw (0.3,-0.1) node[right, align=left,rotate=-50]{{Daher}};
        \draw (1.3,-0.1) node[right, align=left,rotate=-50]{{m{\"u}ssen}};
        \draw (2.3,-0.1) node[right, align=left,rotate=-50]{{wir}};
        \draw (3.3,-0.1) node[right, align=left,rotate=-50]{{die}};
        \draw (4.3,-0.1) node[right, align=left,rotate=-50]{{Voraussetzungen}};
        \draw (5.3,-0.1) node[right, align=left,rotate=-50]{{mit}};
        \draw (6.3,-0.1) node[right, align=left,rotate=-50]{{allergr{\"o}{\ss}ter}};
        \draw (7.3,-0.1) node[right, align=left,rotate=-50]{{Sorgfalt}};
        \draw (8.3,-0.1) node[right, align=left,rotate=-50]{{untersuchen}};
        \draw (9.3,-0.1) node[right, align=left,rotate=-50]{{.}};

        \draw[thick] (0,0) grid (10,13);
        \end{tikzpicture}
  \begin{tikzpicture}[thick,scale=0.35,baseline]
  \small
        \fill[black] (1,1) rectangle (0,0);
        \fill[black] (2,2) rectangle (1,1);
        \fill[black] (7,4) rectangle (6,3);
        \fill[black] (10,3) rectangle (9,2);

        \draw (0.3,-0.1) node[right, align=left,rotate=-50]{{Ich}};
        \draw (1.3,-0.1) node[right, align=left,rotate=-50]{{mu{\ss}}};
        \draw (2.3,-0.1) node[right, align=left,rotate=-50]{{an}};
        \draw (3.3,-0.1) node[right, align=left,rotate=-50]{{dieser}};
        \draw (4.3,-0.1) node[right, align=left,rotate=-50]{{Stelle}};
        \draw (5.3,-0.1) node[right, align=left,rotate=-50]{{das}};
        \draw (6.3,-0.1) node[right, align=left,rotate=-50]{{Haus}};
        \draw (7.3,-0.1) node[right, align=left,rotate=-50]{{dar{\"u}ber}};
        \draw (8.3,-0.1) node[right, align=left,rotate=-50]{{in}};
        \draw (9.3,-0.1) node[right, align=left,rotate=-50]{{Kenntnis}};
        \draw (10.3,-0.1) node[right, align=left,rotate=-50]{{setzen}};

        \draw[thick] (0,0) grid (11,4);
        \end{tikzpicture}
        \label{fig:model}
        \caption{Our Model}
\end{subfigure}
\begin{subfigure}[b]{0.3\textwidth}
\begin{tikzpicture}[thick,scale=0.35,baseline]
\small
        \draw (0.3,-0.1) node[right, align=left,rotate=-50]{{Daher}};
        \draw (1.3,-0.1) node[right, align=left,rotate=-50]{{m{\"u}ssen}};
        \draw (2.3,-0.1) node[right, align=left,rotate=-50]{{wir}};
        \draw (3.3,-0.1) node[right, align=left,rotate=-50]{{die}};
        \draw (4.3,-0.1) node[right, align=left,rotate=-50]{{Voraussetzungen}};
        \draw (5.3,-0.1) node[right, align=left,rotate=-50]{{mit}};
        \draw (6.3,-0.1) node[right, align=left,rotate=-50]{{allergr{\"o}{\ss}ter}};
        \draw (7.3,-0.1) node[right, align=left,rotate=-50]{{Sorgfalt}};
        \draw (8.3,-0.1) node[right, align=left,rotate=-50]{{untersuchen}};
        \draw (9.3,-0.1) node[right, align=left,rotate=-50]{{.}};

        \fill[black] (1,1) rectangle (0,0);
        \fill[black] (2,5) rectangle (1,4);
        \fill[black] (2,4) rectangle (1,3);
        \fill[black] (3,3) rectangle (2,2);
        \fill[black] (4,12) rectangle (3,11);
        \fill[black] (5,12) rectangle (4,11);
        \fill[black] (6,8) rectangle (5,7);
        \fill[black] (7,7) rectangle (6,6);
        \fill[black] (8,8) rectangle (7,7);
        \draw[pattern=north west lines, pattern color=black] (8,6) rectangle (7,5);
        \draw[pattern=north west lines, pattern color=black] (9,9) rectangle (8,8);
        \draw[pattern=north west lines, pattern color=black] (9,10) rectangle (8,9);
        \fill[black] (9,11) rectangle (8,10);
        \fill[black] (10,13) rectangle (9,12);
 \draw[thick] (0,0) grid (10,13);
 \end{tikzpicture}
\begin{tikzpicture}[thick,scale=0.35,baseline]
\small
        \fill[black] (1,1) rectangle (0,0);
        \fill[black] (2,2) rectangle (1,1);
        \fill[black] (10,3) rectangle (9,2);
        \fill[black] (11,3) rectangle (10,2);
        \draw[pattern=north west lines, pattern color=black] (6,4) rectangle (5,3);
        \draw[pattern=north west lines, pattern color=black] (7,4) rectangle (6,3);

        \draw (0.3,-0.1) node[right, align=left,rotate=-50]{{Ich}};
        \draw (1.3,-0.1) node[right, align=left,rotate=-50]{{mu{\ss}}};
        \draw (2.3,-0.1) node[right, align=left,rotate=-50]{{an}};
        \draw (3.3,-0.1) node[right, align=left,rotate=-50]{{dieser}};
        \draw (4.3,-0.1) node[right, align=left,rotate=-50]{{Stelle}};
        \draw (5.3,-0.1) node[right, align=left,rotate=-50]{{das}};
        \draw (6.3,-0.1) node[right, align=left,rotate=-50]{{Haus}};
        \draw (7.3,-0.1) node[right, align=left,rotate=-50]{{dar{\"u}ber}};
        \draw (8.3,-0.1) node[right, align=left,rotate=-50]{{in}};
        \draw (9.3,-0.1) node[right, align=left,rotate=-50]{{Kenntnis}};
        \draw (10.3,-0.1) node[right, align=left,rotate=-50]{{setzen}};
 \draw[thick] (0,0) grid (11,4);
 \end{tikzpicture}
 \caption{Reference}
 \label{fig:ref}
\end{subfigure}
 \caption{Two examples from the German$\to$English alignment test set. Alignments in (a) show the output from \GIZA and (b) from our model (\textit{Multi-task} with full-context and \GIZA supervised).
          Gold Alignments in shown in (c).
          Black squares and hatched squares in the reference represent \textit{sure} and \textit{possible} alignments, respectively.}
 \label{fig:example}
\end{figure*}
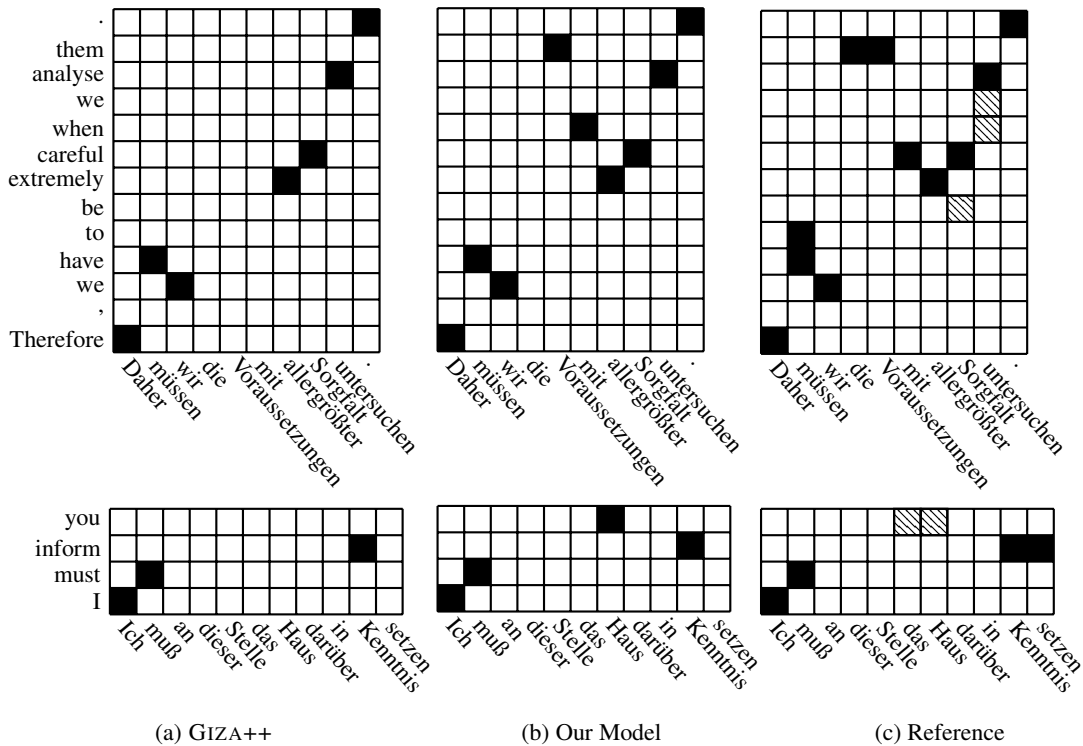

Similar to the trend observed in \ref{sec:resultsAlignment}, providing full target sentence context in the decoder helps the model to improve further and perform comparably to \GIZA. Lastly, supervision with \GIZA gives the best \AER and significantly outperforms \GIZA. The improvements in alignment quality and no degradation in \BLEU compared to the layer average baseline shows the effectiveness of the proposed multi-task approach.

\section{Analysis}
To further investigate why our proposed approach is superior to \GIZA in terms of \AER, 
we analyze the generated word alignments of both models.
We observe that our model tends to align pronouns (e.g.~\textit{you} or \textit{them}) with regular nouns (e.g.~objects or subjects).
Given the gold alignments, it seems that these alignment links are correct or at least possible (\newcite{giza2003} provided annotators two options to specify alignments: \textit{sure} and \textit{possible} for unambiguous and ambiguous alignments respectively).

Figure~\ref{fig:example} shows two examples from the German$\to$English alignment test set.
In the first example, our model correctly aligns \textit{them} with \textit{Voraussetzungen (criteria)}.
The German parliament speaker indeed mentioned \textit{Verfahrensvoraussetzungen (procedural criteria)} in one of the preceding sentences and 
refers later to them by using the term \textit{Voraussetzungen (criteria)}.
In the second example, the pronoun \textit{you} is correctly aligned to the noun \textit{Haus (house)} which is just another way to address the audience in the European parliament.
Both alignment links are not generated by \GIZA.
This could be related to fact that a statistical model is based on counting co-occurrences.
We speculate that to generate such alignment links, a model needs to be able to encode contextual information.
Experimental results in~\cite{tang2018analysis} suggest that NMT models learn to encode contextual information, which seems to be necessary for word sense disambiguation.
Since pronouns can be ambiguous references, we assume that both problems are closely related and therefore believe that the ability to encode contextual information may be beneficial for generating word alignments.

\begin{figure}[!htb]
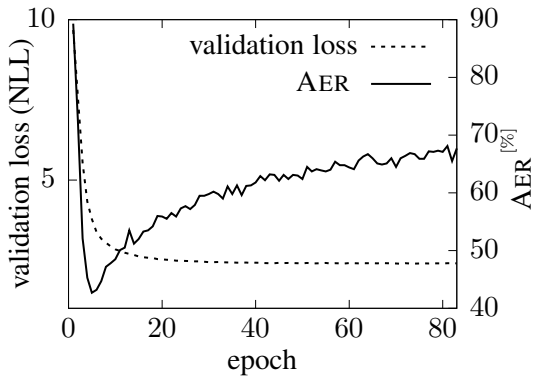

    \include{aerNLL}
    \caption{Test \AER and validation loss (NLL) per epoch on the WMT'18 English$\to$German task.}
    \label{fig:aerNLL}
\end{figure}

From our experiments on the WMT'18 dataset, we observe that the alignment quality of the layer average baseline is quite low (cf.~Table~\ref{tab:resultsAlignTranslate}).
To further investigate this, we plot the test \AER and the validation NLL loss per epoch (Figure~\ref{fig:aerNLL}).
The graph shows that the lowest \AER of 42.7\% is already reached in the fifth epoch. 
This suggests that picking an earlier checkpoint for generating word alignments could be beneficial for better supervision.
Unfortunately, an alignment validation set does not exist for this task.
\section{Related Work}
Leveraging alignments obtained from statistical MT toolkits to guide NMT attention mechanisms has been explored in the past. 
\newcite{Mi2016}, \newcite{chen2016guided}, \newcite{Liu2016} and \newcite{alkhouli17:alignbiasattention} supervise the attention mechanisms of recurrent models~\cite{bahdanau2014neural} in this way.
Our multi-task framework is inspired by these publications. However, we examine its effect on the Transformer model
~\cite{vaswani2017transformer}, which provides state-of-the-art results on several translation benchmarks. Previous works report 
significant gains in translation accuracy in low resource settings, however gains remain modest given larger amounts of parallel 
data (millions of sentences). These approaches also fail to achieve significantly better alignment accuracy than the statistical MT toolkits.
\newcite{peter2017:Foresight} and \newcite{li2018target} improve upon the previous works in terms of alignment accuracy by 
providing an alignment module with additional information about the to-be-aligned target word. Expanding on this idea,
we propose to leverage the full target sentence context leading to \AER improvements.
\newcite{Zenkel2019} presents an approach that eliminates the reliance on statistical word aligners by instead by directly optimizing the attention activations for predicting the target word. We empirically compare our approach of obtaining high quality alignments without the need of statistical word aligners to \newcite{Zenkel2019}.

Augmenting the task objective with linguistic information, such as word alignments, also has had applications beyond MT.
\newcite{Strubell2018} showed that adding linguistic information from parse trees into one of the attention heads of the transformer model can help in the semantic role labeling.
Inspired by \newcite{Strubell2018}, we inject the alignment information through one of the attention heads for the translation task instead. 

As a by-product of developing our model, we present a simple way to quantitatively evaluate and analyze the quality of attention probabilities learnt by 
different parts of the Transformer model with respect to modeling alignments, which contributes to previous work on understanding attention mechanisms ~\cite{ghader2017does, raganato2018analysis, tang2018analysis}.

\section{Conclusions}
This paper addresses the task of jointly learning to produce translations and alignments with a single Transformer model. 
By using a multi-task objective along with providing full target sentence context to our alignment module, we are able to produce
better alignments than previous approaches not relying on external alignment toolkits. We demonstrate that our framework can be 
extended to use external alignments from \GIZA to achieve significantly better alignment results compared to \GIZA, while 
maintaining the same translation performance.

Currently, our self-training based approach needs two training runs.
To train our model in a single run, we would like to investigate a training method which alternates between alignment extraction and model training. 
    
\section*{Acknowledgments}
We would like to thank the LILT team for releasing scripts and datasets for alignment evaluation. We are also grateful to Andrew Finch, Matthias Sperber, Barry Theobald and the anonymous reviewers for their helpful comments. Many thanks to Dorothea Peitz for helpful discussions about significance testing, Yi-Hsiu Liao for suggesting interesting extensions and the rest of Siri Machine Translation Team for their support.

\bibliography{aa}
\bibliographystyle{acl_natbib}
\end{document}

%% file: aerNLL.tex
\begingroup
  \makeatletter
  \providecommand\color[2][]{%
    \GenericError{(gnuplot) \space\space\space\@spaces}{%
      Package color not loaded in conjunction with
      terminal option `colourtext'%
    }{See the gnuplot documentation for explanation.%
    }{Either use 'blacktext' in gnuplot or load the package
      color.sty in LaTeX.}%
    \renewcommand\color[2][]{}%
  }%
  \providecommand\includegraphics[2][]{%
    \GenericError{(gnuplot) \space\space\space\@spaces}{%
      Package graphicx or graphics not loaded%
    }{See the gnuplot documentation for explanation.%
    }{The gnuplot epslatex terminal needs graphicx.sty or graphics.sty.}%
    \renewcommand\includegraphics[2][]{}%
  }%
  \providecommand\rotatebox[2]{#2}%
  \@ifundefined{ifGPcolor}{%
    \newif\ifGPcolor
    \GPcolorfalse
  }{}%
  \@ifundefined{ifGPblacktext}{%
    \newif\ifGPblacktext
    \GPblacktexttrue
  }{}%
  \let\gplgaddtomacro\g@addto@macro
  \gdef\gplbacktext{}%
  \gdef\gplfronttext{}%
  \makeatother
  \ifGPblacktext
    \def\colorrgb#1{}%
    \def\colorgray#1{}%
  \else
    \ifGPcolor
      \def\colorrgb#1{\color[rgb]{#1}}%
      \def\colorgray#1{\color[gray]{#1}}%
      \expandafter\def\csname LTw\endcsname{\color{white}}%
      \expandafter\def\csname LTb\endcsname{\color{black}}%
      \expandafter\def\csname LTa\endcsname{\color{black}}%
      \expandafter\def\csname LT0\endcsname{\color[rgb]{1,0,0}}%
      \expandafter\def\csname LT1\endcsname{\color[rgb]{0,1,0}}%
      \expandafter\def\csname LT2\endcsname{\color[rgb]{0,0,1}}%
      \expandafter\def\csname LT3\endcsname{\color[rgb]{1,0,1}}%
      \expandafter\def\csname LT4\endcsname{\color[rgb]{0,1,1}}%
      \expandafter\def\csname LT5\endcsname{\color[rgb]{1,1,0}}%
      \expandafter\def\csname LT6\endcsname{\color[rgb]{0,0,0}}%
      \expandafter\def\csname LT7\endcsname{\color[rgb]{1,0.3,0}}%
      \expandafter\def\csname LT8\endcsname{\color[rgb]{0.5,0.5,0.5}}%
    \else
      \def\colorrgb#1{\color{black}}%
      \def\colorgray#1{\color[gray]{#1}}%
      \expandafter\def\csname LTw\endcsname{\color{white}}%
      \expandafter\def\csname LTb\endcsname{\color{black}}%
      \expandafter\def\csname LTa\endcsname{\color{black}}%
      \expandafter\def\csname LT0\endcsname{\color{black}}%
      \expandafter\def\csname LT1\endcsname{\color{black}}%
      \expandafter\def\csname LT2\endcsname{\color{black}}%
      \expandafter\def\csname LT3\endcsname{\color{black}}%
      \expandafter\def\csname LT4\endcsname{\color{black}}%
      \expandafter\def\csname LT5\endcsname{\color{black}}%
      \expandafter\def\csname LT6\endcsname{\color{black}}%
      \expandafter\def\csname LT7\endcsname{\color{black}}%
      \expandafter\def\csname LT8\endcsname{\color{black}}%
    \fi
  \fi
    \setlength{\unitlength}{0.0500bp}%
    \ifx\gptboxheight\undefined%
      \newlength{\gptboxheight}%
      \newlength{\gptboxwidth}%
      \newsavebox{\gptboxtext}%
    \fi%
    \setlength{\fboxrule}{0.5pt}%
    \setlength{\fboxsep}{1pt}%
\begin{picture}(4320.00,2590.00)%
    \gplgaddtomacro\gplbacktext{%
      \csname LTb\endcsname
      \put(496,1223){\makebox(0,0)[r]{\strut{}$5$}}%
      \put(496,2429){\makebox(0,0)[r]{\strut{}$10$}}%
      \put(592,99){\makebox(0,0){\strut{}$0$}}%
      \put(1290,99){\makebox(0,0){\strut{}$20$}}%
      \put(1987,99){\makebox(0,0){\strut{}$40$}}%
      \put(2685,99){\makebox(0,0){\strut{}$60$}}%
      \put(3382,99){\makebox(0,0){\strut{}$80$}}%
      \put(3583,259){\makebox(0,0)[l]{\strut{}$40$}}%
      \put(3583,693){\makebox(0,0)[l]{\strut{}$50$}}%
      \put(3583,1127){\makebox(0,0)[l]{\strut{}$60$}}%
      \put(3583,1561){\makebox(0,0)[l]{\strut{}$70$}}%
      \put(3583,1995){\makebox(0,0)[l]{\strut{}$80$}}%
      \put(3583,2429){\makebox(0,0)[l]{\strut{}$90$}}%
    }%
    \gplgaddtomacro\gplfronttext{%
      \csname LTb\endcsname
      \put(240,1344){\rotatebox{-270}{\makebox(0,0){\strut{}validation loss (NLL)}}}%
      \put(3967,1344){\rotatebox{-270}{\makebox(0,0){\strut{}\AERpercent}}}%
      \put(2039,-141){\makebox(0,0){\strut{}epoch}}%
      \csname LTb\endcsname
      \put(2752,2226){\makebox(0,0)[r]{\strut{}validation loss}}%
      \csname LTb\endcsname
      \put(2752,1946){\makebox(0,0)[r]{\strut{}\AER}}%
    }%
    \gplbacktext
    \put(0,0){\includegraphics{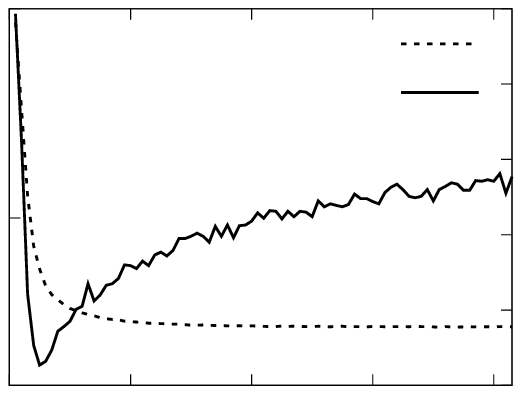}}%
    \gplfronttext
  \end{picture}%
\endgroup